\documentclass[11pt,a4paper]{article}
\usepackage{naaclhlt2019}
\usepackage{times}
\usepackage{latexsym}
\usepackage{graphicx}
\usepackage{url}
\usepackage{subcaption}

\aclfinalcopy 
\def\aclpaperid{495} 

\title{Detecting Dementia in Mandarin Chinese using Transfer Learning from a Parallel Corpus}

\author{Bai Li$^{1,2}$, Yi-Te Hsu$^{1,2,4}$, Frank Rudzicz$^{1,2,3}$ \\
    $^1$ University of Toronto, Toronto, Canada \\
    $^2$ Vector Institute, Toronto, Canada \\
    $^3$ Toronto Rehabilitation Institute, Toronto, Canada \\
    $^4$ Academia Sinica, Taipei, Taiwan \\
  {\tt \{bai, eeder, frank\}@cs.toronto.edu} 
  } 
\date{}

\begin{document}
\maketitle
\begin{abstract}
Machine learning has shown promise for automatic detection of Alzheimer's disease (AD) through speech; however, efforts are hampered by a scarcity of data, especially in languages other than English. 
We propose a method to learn a correspondence between independently engineered lexicosyntactic features in two languages, using a large parallel corpus of out-of-domain movie dialogue data. 
We apply it to dementia detection in Mandarin Chinese, and demonstrate that our method outperforms both unilingual and machine translation-based baselines.
This appears to be the first study that transfers feature domains in detecting cognitive decline. 
\end{abstract}

\section{Introduction}

\begin{figure*}[ht]
\begin{center}
  \includegraphics[width=\linewidth]{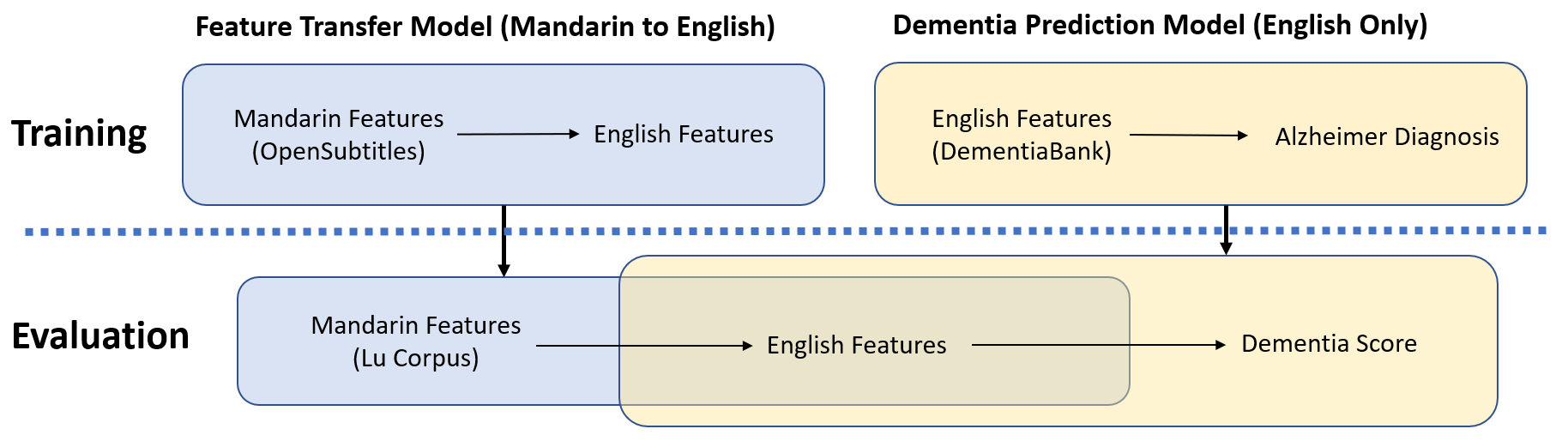}
  \caption{Diagram of our model. We train two separate models: the first is trained on OpenSubtitles and learns to map Mandarin features to English features; the second is trained on DementiaBank and predicts dementia given English features. During evaluation, the two models are combined to predict dementia in Mandarin.}
  \label{fig:model-diagram}
\end{center}
\end{figure*}

Alzheimer's disease (AD) is a neurodegenerative disease affecting 5.7 million people in the US \citep{alzheimer-is-bad}, and is the most common cause of dementia. Although no cure yet exists, early detection of AD is crucial for an effective treatment to delay or prepare for its effects \citep{alzheimer-detection-is-good}. One of the earliest symptoms of AD is speech impairment, including a difficulty in finding words and changes to grammatical structure \citep{alzheimer-language-impairment}. These early signs can be detected by having the patient perform a picture description task, such as the {\em Cookie Theft} task from the Boston Diagnostic Aphasia Examination \citep{boston-diagnostic}.

Previous models have applied machine learning to automatic detection of AD, for example, \citet{fraser2016} extracted a wide variety of lexicosyntactic and acoustic features to classify AD and obtained 82\% accuracy on the DementiaBank (DB) dataset. However, clinical studies of AD are expensive, so datasets of patient data are often scarce. \citet{chloe-normative} augmented DB with more a much larger corpus of normative data and improved the classification accuracy to 93\% on DB. Similar linguistic differences between healthy and AD speech have been observed in Mandarin Chinese \citep{lai2009}, but machine learning has not yet been applied to detecting AD in Mandarin.

\citet{daume-frustratingly-easy} proposed a simple way of combining features in different domains, assuming that the same features are extracted in each domain. In our case, ensuring consistency of features across domains is challenging because of the grammatical differences between Mandarin and English. For example, Mandarin doesn't have determiners or verb tenses, and has classifiers, which don't exist in English \citep{chinese-grammar}. Another method trains a classifier jointly on multiple domains with different features on each domain, by learning a projection to a common subspace \citep{duan2012}. However, this method only accepts labelled samples in each domain, and cannot make use of unlabelled, out-of-domain data. Other work from our broader group \citep{fraser-information-units} combined English and French data by extracting features based on conceptual ``information units'' rather than words, thus limiting the effects of multilingual differences.

In the current work, we train an {\em unsupervised} model to detect dementia in Mandarin, requiring only the English DB dataset and a large parallel Mandarin-English corpus of normative dialogue. 
We extract lexicosyntactic features in Mandarin and English using separate pipelines, and use the OpenSubtitles corpus of bilingual parallel movie dialogues to learn a correspondence between the different feature sets. We combine this correspondence model with a classifier trained on DB to predict dementia on Mandarin speech. To evaluate our system, we apply it to a dataset of speech from Mandarin-speakers with dementia, and demonstrate that our method outperforms several baselines.

\section{Datasets}

We use the following datasets:

\begin{itemize}
    \item {\bf DementiaBank} \citep{dementiabank}: a corpus of  {\em Cookie Theft} picture descriptions, containing 241 narrations from healthy controls and 310 from patients with dementia. Each narration is professionally transcribed and labelled with part-of-speech tags. In this work, we only use the narration transcripts, and neither the part-of-speech tags or raw acoustics.
    \item {\bf Lu Corpus} \citep{aphasiabank}: contains 49 patients performing the {\em Cookie theft} picture description, category fluency, and picture naming tasks in Taiwanese Mandarin. The picture description narrations were human-transcribed; patient diagnoses are unspecified but exhibit various degrees of dementia.
    \item {\bf OpenSubtitles2016} \citep{opensubtitles}: a corpus of parallel dialogues extracted from movie subtitles in various languages. We use the Traditional Chinese / English language pair, which contains 3.3 million lines of dialogue.
\end{itemize}

The Lu Corpus is missing specifics of diagnosis, so we derive a {\em dementia score} for each patient using the category fluency and picture naming tasks. For each category fluency task, we count the number of unique items named; for the picture naming tasks, we score the number of pictures correctly named, awarding partial credit if a hint was given. We apply PCA to the scores across all tasks, and assign the first principal component to be the {\em dementia score} for each patient. This gives a relative ordering of all patients for degree of dementia, which we treat as the ground-truth for evaluating our models.

\section{Methodology}

\subsection{Feature Extraction}

We extract a variety of lexicosyntactic features in Mandarin and English, including type-token-ratio, the number of words per sentence, and proportions of various part-of-speech tags\footnote{The feature extraction pipeline is open-source, available at: \url{https://github.com/SPOClab-ca/COVFEFE}. The {\tt lex} and {\tt lex\_chinese} pipelines were used for English and Chinese, respectively.}. A detailed description of the features is provided in the supplementary materials (Section \ref{sec:feature-extraction-appendix}). In total, we extract 143 features in Mandarin and 185 in English. To reduce sparsity, we remove features in both languages that are constant for more than half of the dataset.

Due to the size of the OpenSubtitles corpus, it was computationally unfeasible to run feature extraction on the entire corpus. Therefore, we randomly select 50,000 narrations from the corpus, where each narration consists of between 1 to 50 contiguous lines of dialogue (about the length of a {\em Cookie Theft} narration).

\label{sec:english-logistic-regression}

For English, we train a logistic regression classifier to classify between dementia and healthy controls on DB, using our features as input. Using L1 regularization and 5-fold CV, our model achieves 77\% classification accuracy on DB. This is slightly lower than the 82\% accuracy reported by \citet{fraser2016}, but it does not include any acoustic features as input.

\subsection{Feature Transfer}

Next, we use the OpenSubtitles corpus to train a model to transform Mandarin feature vectors to English feature vectors. For each target English feature, we train a separate ElasticNet linear regression \citep{elasticnet}, using the Mandarin features of the parallel text as input. We perform a hyperparameter search independently for each target feature, using 3-fold CV to minimize the MSE.

\subsection{Regularization}

Although the output of the ElasticNet regressions may be given directly to the logistic regression model to predict dementia, this method has two limitations. First, the model considers each target feature separately and cannot take advantage of correlations between target features. Second, it treats all target feature equally, even though some are noisier than others. We introduce two regularization mechanisms to address these drawbacks: reduced rank regression and joint feature selection.

\subsubsection*{Reduced Rank Regression} \label{sec:reduced-rank}

Reduced rank regression (RRR) trains a single linear model to predict all the target features: it minimizes the sum of MSE across all target features, with the constraint that the rank of the linear mapping is bounded by some given $R$ \citep{reduced-rank}. Following recommended procedures \citep{procedures-rrr}, we standardize the target features and find the best value of $R$ with cross validation. However, this procedure did not significantly improve results so it was not included in our best model.

\subsubsection*{Joint Feature Selection} \label{sec:feature-selection}

A limitation of the above models is that they are not robust to noisy features. For example, if some English feature is useful for predicting dementia, but cannot be accurately predicted using the Mandarin features, then including this feature might hurt the overall performance. A desirable English feature in our pipeline needs to not only be useful for predicting dementia in English, but also be reconstructable from Mandarin features.

\begin{figure*}[t]
\centering
  \includegraphics[height=7cm]{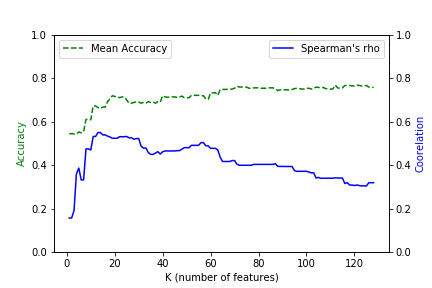}
  \caption{Accuracy of DementiaBank classifier model and Spearman's $\rho$ on Lu corpus, using only the top $K$ English features ordered by $R^2$ on the OpenSubtitles corpus. Spearman's $\rho$ is maximized at $K = 13$, achieving a score of $\rho = 0.549$. DementiaBank accuracy generally increases with more features.}
  \label{fig:top-k-plot}
\end{figure*}

We modify our pipeline as follows. After training the ElasticNet regressions, we sort the target features by their $R^2$ (coefficient of determination) measured on the training set, where higher values indicate a better fit. Then, for each $K$ between 1 and the number of features, we select only the top $K$ features and re-train the DB classifier (\ref{sec:english-logistic-regression}) to only use those features as input. The result of this experiment is shown in Figure \ref{fig:top-k-plot}.

\section{Experiments}

\subsection{Baseline Models}

We compare our system against two simple baselines:

\begin{enumerate}
    \item {\bf Unilingual baseline:} using the Mandarin features, we train a linear regression to predict the dementia score. We take the mean across 5 cross-validation folds.
    \item {\bf Translate baseline:} The other intuitive way to generate English features from a Mandarin corpus is by using translation. We use Google Translate\footnote{\url{https://translate.google.com/}} to translate each Mandarin transcript to English. Then, we extract features from the translated English text and feed them to the dementia classifier described in section \ref{sec:english-logistic-regression}.
\end{enumerate}

\subsection{Evaluation Metric}

We evaluate each model by comparing the Spearman's rank-order correlation $\rho$ \citep{spearman-rho} between the ground truth dementia scores and the model's predictions. This measures the model's ability to rank the patients from the highest to the lowest severities of dementia, without requiring a threshold value.

\subsection{Experimental Results}

\begin{table}[h]
\begin{center}
 \begin{tabular}{||l l||} 
 \hline
 {\bf Model} & {\bf Spearman $\rho$} \\
 \hline\hline
 {\bf Baselines} & \\
 \ \ \ \ Unilingual & 0.385 \\
 \ \ \ \ Google Translate & 0.366 \\ \hline
 {\bf Our models} & \\
 \ \ \ \ Feature Transfer & 0.319 \\
 \ \ \ \ + RRR & 0.354 \\
 \ \ \ \ + JFS & {\bf 0.549} \\ \hline
\end{tabular}
\end{center}
\caption{\label{results-table} Baselines compared with our models, evaluated on the Lu corpus. RRR: Reduced rank regression (\ref{sec:reduced-rank}), JFS: Joint feature selection (\ref{sec:feature-selection}).}
\end{table}

Our best model achieves a Spearman's $\rho$ of 0.549, beating the translate baseline ($n$ = 49, $p$ = 0.06). Joint feature selection appears to be crucial, since the model performs worse than the baselines if we use all of the features. This is the case no matter if we predict each target feature independently or all at once with reduced rank regression. RRR does not outperform the baseline model, probably because it fails to account for the noisy target features in the correspondence model and considers each feature equally important. We did not attempt to use joint feature selection and RRR at the same time, because the multiplicative combination of hyperparameters $K$ and $R$ would produce a multiple comparisons problem using the small validation set.

Using joint feature selection, we find that the best score is achieved when we use $K = 13$ target features (Figure \ref{fig:top-k-plot}). With $K < 13$, performance suffers because the DementiaBank classifier is not given enough information to make accurate classifications. With $K > 13$, the accuracy for the DementiaBank classifier improves; however, the overall performance degrades because it is given noisy features with low $R^2$ coefficients. A list of the top features is given in Table \ref{table:top-joint-features} in the supplementary materials.

In our experiments, the correspondence model worked better when absolute counts were used for the Chinese CFG features (e.g., the number of $NP \to PN$ productions in the narration) rather than ratio features (e.g., the proportion of CFG productions that were $NP \to PN$). When ratios were used for source features, the $R^2$ coefficients for many target features decreased. A possible explanation is that the narrations have varying lengths, and dividing features by the length introduces a nonlinearity that adversely affects our linear models. However, more experimentation is required to examine this hypothesis.


\subsection{Ablation Study}

\begin{figure}
  \centering
  \includegraphics[width=\linewidth]{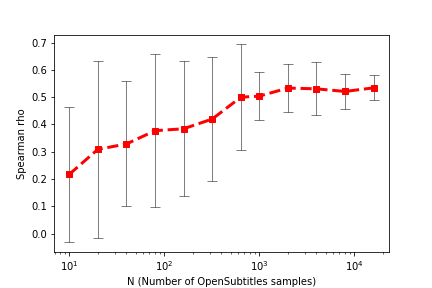}
  \caption{Ablation experiment where a various number of OpenSubtitles samples were used for training. The error bars indicate the two standard deviation confidence interval.}
  \label{fig:ablation}
\end{figure}

Next, we investigate how many parallel OpenSubtitles narrations were necessary to learn the correspondence model. We choose various training sample sizes from 10 to 50,000 and, for each training size, we train and evaluate the whole model from end-to-end 10 times with different random seeds (Figure \ref{fig:ablation}). As expected, the Spearman's $\rho$ increased as more samples were used, but only 1000-2000 samples were required to achieve comparable performance to the full model.

\section{Conclusion}

We propose a novel method to use a large parallel corpus to learn mappings between engineered features in two languages. Combined with a dementia classifier model for English speech, we constructed a model to predict dementia in Mandarin Chinese. Our method achieves state-of-the-art results for this task and beats baselines based on unilingual models and Google Translate. It is successful despite the stark differences between English and Mandarin, and the fact that the parallel corpus is out-of-domain for the task. Lastly, our method does not require any Mandarin data for training, which is important given the difficulty of acquiring sensitive clinical data.

Future work will investigate the use of automatic speech recognition to reduce the need for manual transcripts, which are impractical in a clinical setting. Also, our model only uses lexicosyntactic features, and ignores acoustic features (e.g., pause duration) which are significant for dementia detection in English. Finally, it remains to apply this method to other languages, such as French \citep{fraser-information-units}, for which datasets have recently been collected.

\section*{Acknowledgements}

We thank Kathleen Fraser and Nicklas Linz for their helpful comments and earlier collaboration which inspired this project.

\bibliography{naaclhlt2019}
\bibliographystyle{acl_natbib}

\appendix

\section{Appendices}

\subsection{Description of Lexicosyntactic Features}
\label{sec:feature-extraction-appendix}

We extract 185 lexicosyntactic features in English and 143 in Mandarin Chinese. We use Stanford CoreNLP to do constituency parsing and part-of-speech tagging \citep{corenlp-english, corenlp-chinese}. We also use {\tt wordfreq} \citep{wordfreq} for word frequency statistics in both languages. Our features are similar to the set of features used by \citet{fraser2016}, which the reader can refer to for a more thorough description.

The following features are extracted in English:

\begin{itemize}
\item {\bf Narrative length:} Number of words and sentences in narration.
\item {\bf Vocabulary richness:} Type-token ratio, moving average type-token ratio (with window sizes of 10, 20, 30, 40, and 50 words), Honor\'e's statistic, and Brun\'et's index.
\item {\bf Frequency metrics:} Mean word frequencies for all words, nouns, and verbs.
\item {\bf POS counts:} Counts and ratios of nouns, verbs, inflected verbs, determiners, demonstratives, adjectives, adverbs, function words, interjections, subordinate conjunctions, and coordinate conjunctions. Also includes some special ratios such as pronoun / noun and noun / verb ratios.
\item {\bf Syntactic complexity:} Counts and mean lengths of clauses, T-units, dependent clauses, and coordinate phrases as computed by Lu's syntactic complexity analyzer \citep{lu-syntactic-analyzer}.
\item {\bf Tree statistics:} Max, median, and mean heights of all CFG parse trees in the narration.
\item {\bf CFG ratios:} Ratio of CFG production rule count for each of the 100 most common CFG productions from the constituency parse tree.
\end{itemize}

The following features are extracted in Mandarin Chinese:

\begin{itemize}
\item {\bf Narrative length:} Number of sentences, number of characters, and mean sentence length.
\item {\bf Frequency metrics:} Type-token ratio, mean and median word frequencies.
\item {\bf POS counts:} For each part-of-speech category, the number of it in the utterance and ratio of it divided by the number of tokens. Also includes some special ratios such as pronoun / noun and noun / verb ratios.
\item {\bf Tree statistics:} Max, median, and mean heights of all CFG parse trees in the narration.
\item {\bf CFG counts:} Number of occurrences for each of the 60 most common CFG production rules from the constituency parse tree.
\end{itemize}
  
\subsection{Top Joint Features}

Table \ref{table:top-joint-features} lists the top English features for joint feature selection (most reconstructable from Chinese features), ordered by $R^2$ coefficients on the OpenSubtitles corpus. The top performing model uses the first 13 features.

\begin{table*}
    \centering
    \footnotesize
    \begin{tabular}{|r l|l|}
    \hline
    \# & {\bf Feature Name} & $R^2$ \\ \hline
    1 & Number of words & 0.894\\
    2 & Number of sentences & 0.828\\
    3 & Brun\'et's index & 0.813\\
    4 & Type token ratio & 0.668\\
    5 & Moving average TTR (50 word window) & 0.503\\
    6 & Moving average TTR (40 word window) & 0.461\\
    7 & Moving average TTR (30 word window) & 0.411\\
    8 & Average word length & 0.401\\
    9 & Moving average TTR (20 word window) & 0.360\\
    10 & Moving average TTR (10 word window) & 0.328\\
    11 & NP $\to$ PRP & 0.294\\
    12 & Number of nouns & 0.233\\
    13 & Mean length of clause & 0.225\\ \hline
    14 & PP $\to$ IN NP & 0.224\\
    15 & Total length of PP & 0.222\\
    16 & Complex nominals per clause & 0.220\\
    17 & Noun ratio & 0.213\\
    18 & Pronoun ratio & 0.208\\
    19 & Number of T-units & 0.207\\
    20 & Number of PP & 0.205\\
    21 & Number of function words & 0.198\\
    22 & Subordinate / coordinate clauses & 0.193\\
    23 & Mean word frequency & 0.193\\
    24 & Number of pronouns & 0.191\\
    25 & Average NP length & 0.188\\

    \hline
    \end{tabular}
    \caption{Top English features for joint feature selection.}
    \label{table:top-joint-features}
\end{table*}

\end{document}